\documentclass[10pt]{article} 
\usepackage[preprint]{tmlr}

\usepackage{amsmath,amsfonts,bm}

\def\eqref#1{equation~\ref{#1}}

\def\1{\bm{1}}

\DeclareMathAlphabet{\mathsfit}{\encodingdefault}{\sfdefault}{m}{sl}
\SetMathAlphabet{\mathsfit}{bold}{\encodingdefault}{\sfdefault}{bx}{n}

\DeclareMathOperator*{\argmin}{arg\,min}

\usepackage{hyperref}
\usepackage{url}
\usepackage{microtype}
\usepackage{graphicx}
\usepackage{subfigure}
\usepackage{booktabs} 
\usepackage{bbding}
\usepackage{graphicx}
\usepackage{tablefootnote}
\usepackage{hyperref}
\usepackage{url}
\usepackage{pifont}
\usepackage{fancyhdr}
\usepackage{multirow}
\usepackage{wrapfig}
\usepackage{bbold}
\usepackage{amsmath,amssymb}
\usepackage{makecell}
\usepackage{amsmath}
\usepackage{amssymb}
\usepackage{mathtools}
\usepackage{amsthm}
\usepackage{caption}
\usepackage[capitalize,noabbrev]{cleveref}
\theoremstyle{plain}

\theoremstyle{definition}

\theoremstyle{remark}

\title{Trimming Down Large Spiking Vision Transformers via Heterogeneous Quantization Search}

\author{\name Boxun Xu$^\ddag$ \email boxunxu@ucsb.edu \\
        \addr Department of Electrical and Computer Engineering\\
      University of California, Santa Barbara
      \AND
      \name Yufei Song$^\ddag$ \email yufei\_song@ucsb.edu \\
      \addr Department of Electrical and Computer Engineering\\
      University of California, Santa Barbara
      \AND
      \name Peng Li$^*$ \email lip@ucsb.edu \\
      \addr Department of Electrical and Computer Engineering\\
      University of California, Santa Barbara
    }

\newcommand{\method}{${SpikeHQ}$}

\def\month{MM}  
\def\year{YYYY} 
\def\openreview{\url{https://openreview.net/forum?id=XXXX}} 

\begin{document}

\maketitle
\def\thefootnote{$\ddag$}\footnotetext{Equal contribution.}
\def\thefootnote{$*$}\footnotetext{Corresponding author.}

\begin{abstract}
Spiking Neural Networks (SNNs) are amenable to deployment on edge devices and neuromorphic hardware due to their lower dissipation. Recently, SNN-based transformers have garnered significant interest, incorporating attention mechanisms akin to their counterparts in Artificial Neural Networks (ANNs) while demonstrating excellent performance. However, deploying large spiking transformer models on resource-constrained edge devices such as mobile phones, still poses significant challenges resulted from the high computational demands of large uncompressed high-precision models. In this work, we introduce a novel heterogeneous quantization method for compressing spiking transformers through layer-wise quantization. Our approach optimizes the quantization of each layer using one of two distinct quantization schemes, i.e., uniform or power-of-two quantification, with mixed bit resolutions. Our heterogeneous quantization demonstrates the feasibility of maintaining high performance for spiking transformers while utilizing an average effective resolution of  $3.14$-$3.67$ bits with less than a 1\% accuracy drop on DVS Gesture and CIFAR10-DVS datasets. It attains a model compression rate of $8.71\times$-$10.19\times$ for standard floating-point spiking transformers. Moreover, the proposed approach achieves a significant energy reduction of $5.69\times$, $8.72\times$, and $10.2\times$ while maintaining high accuracy levels of $85.3\%$, $97.57\%$,  and $80.4\%$ on the N-Caltech101, DVS-Gesture, and CIFAR10-DVS datasets, respectively.

\end{abstract}

\section{Introduction}
\label{Introduction}

\begin{figure}[h]
    \centering
    \includegraphics[width=\columnwidth, clip, trim={0cm 3cm 2cm 5cm}]{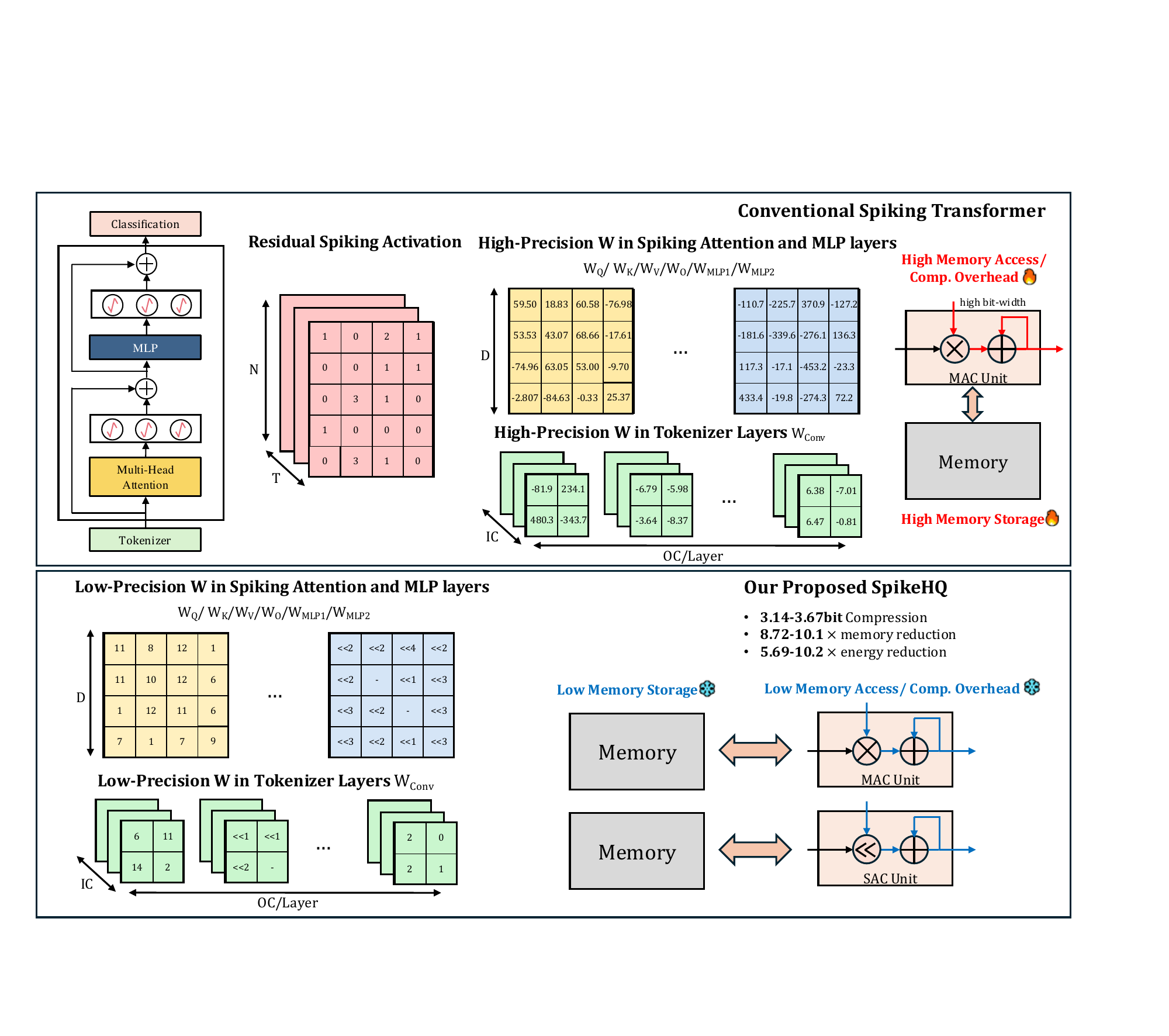}
    \caption{Heterogeneous Quantization Compression on Spiking Vision Transformers.}
    \label{fig:spikformerarch}
\end{figure}

Vision transformers (ViTs) have emerged as a powerful vision backbone substitute to convolutional neural networks(CNNs) \citep{Attention, ViT}. However, ViTs' state-of-the-art accuracy comes at the price of prohibitive hardware latency and energy consumption for both training on the cloud and inference on the edge, limiting their prevailing deployment on resource-constrained devices\citep{Ecoformer, Adder_attention}. Spiking neural networks (SNNs), as the third generation of neural networks, are models of computation that more closely resemble biological neurons with real-time processing and low-power dissipation. SNNs can be seamlessly adapted to edge devices, enabled by great progress in neuromorphic applications \citep{Loihi, TrueNorth} with high accuracy\citep{TSSL-BP}. More recently, several works have demonstrated the potential of realizing spiking transformers with low-power dissipation to address the aforementioned issues associated with ViTs. \citep{spikformer} propose a spiking-based self-attention mechanism and adapt the existing transformer architecture into spiking neural networks for vision tasks, reducing energy consumption while delivering improved performance over CNN-based SNNs. \citep{xu2023dista} further improve spiking neurons' spatiotemporal attention abilities and enhance attention map's representativeness by denoising functions or hashmaps\citep{xu2024ds2ta}. In addition, as a variant of SNNs, the sparse and event-driven nature of spikes causes the spiking ViT's computation to leverage multiple neuromorphic hardware platforms \citep{davies2018loihi, debole2019truenorth, PTB, xu2024spiking} to support a real-time and low-power computation paradigm. A typical spiking transformer architecture is depicted in Fig.~\ref{fig:spikformerarch}. 


Despite the encouraging progress on SNN-based ViTs, one pressing hurdle is that these approaches still suffer from large model sizes and expensive use of full-precision weights and multiplication operations. As such, these models are not amenable to deployment on resource-strained edge devices due to prohibitively high computational/memory access overheads, and energy consumption. 

Quantization is inherently hardware-friendly and offers a promising model compression solution, and has been adopted for DNN/SNN hardware accelerators. \citep{IJCNN_QSNN_qf} proposes a uniform and non-layer-wise quantization framework Spiking CNNs. \citep{Tang23} binarizes the weight of all layers and replaces multiplication with XOR operations to in spiking CNNs. However, precision loss in uniform quantization and unsystematic aggressive quantization can lead to large performance drop and limit model compression ratio. Despite the promise of quantization for model compression, the use of quantization for SNNs and transformers in general has been limited. 
We introduce {\method}, the first approach to heterogeneous quantization for spiking neural networks, with a specific focus on spiking transformers. With a neural architecture search formulation, {\method} employs optimal layer-by-layer compression, utilizing either a uniform or power-of-two quantization scheme with mixed bit resolutions. While quantization has been employed for both artificial neural network (ANN) and spiking neural network (SNN) accelerators, heterogeneous quantization has not been explored for SNNs in general and ANN-based transformers.

Our experimental studies reveal that {\method} can compress the weights in a given spiking transformer from 32-bit floating-point numbers to 2-bit integers, achieving an impressive compression rate of $16\times$. {\method}  showcases the feasibility of maintaining high performance for spiking transformers with an average effective resolution of $2.33$-$3.24$ bits, and less than a 1\% accuracy drop on DVS Gesture and CIFAR10-DVS datasets. {\method} demonstrates a substantial reduction in total energy consumption, ranging from $5.69\times$ to $10.2\times$, and a storage overhead reduction between $8.71\times$ and $10.19\times$, all while maintaining competitive task performance when compared to the high-precision unquantized spiking transformers. Furthermore, our empirical findings underscore specific layers/computations within a spiking transformer that necessitate high-precision weight parameters. These layers are identified as key bottlenecks for potential efficiency improvements, highlighting the need for targeted enhancements in future architectural and design innovations.

\section{Background}
\subsection{Spiking Transformers}
\citep{spikformer} adapts self-attention mechanisms \citep{Attention} and ANN-based transformer architectures for spiking neural network (SNN) based implementation to improve learning and efficiency. It takes a 2D image sequence as input and uses the Spiking tokenizer module to project the image into a spike-form feature vector, which is further divided into flattened spike-form patches. A spike-form relative position embedding is generated and added to the patches. The architecture proposed in \citep{spikformer} includes L-block Spikformer encoders, each consisting of a Spiking Self Attention (SSA) module and an MLP block. Residual connections are established within the SSA and MLP blocks. The SSA module  models local-global information using spike-form Query, Key, and Value components without employing softmax. A global average-pooling  operation is performed on the processed features, and the resulting feature vector is fed into a fully-connected classification head to obtain the final prediction. 

\subsection{Heterogeneous Quantization}
\begin{figure*}
    \centering
    \includegraphics[width=\textwidth, clip, trim={3cm 1.5cm 4.2cm 1cm}]{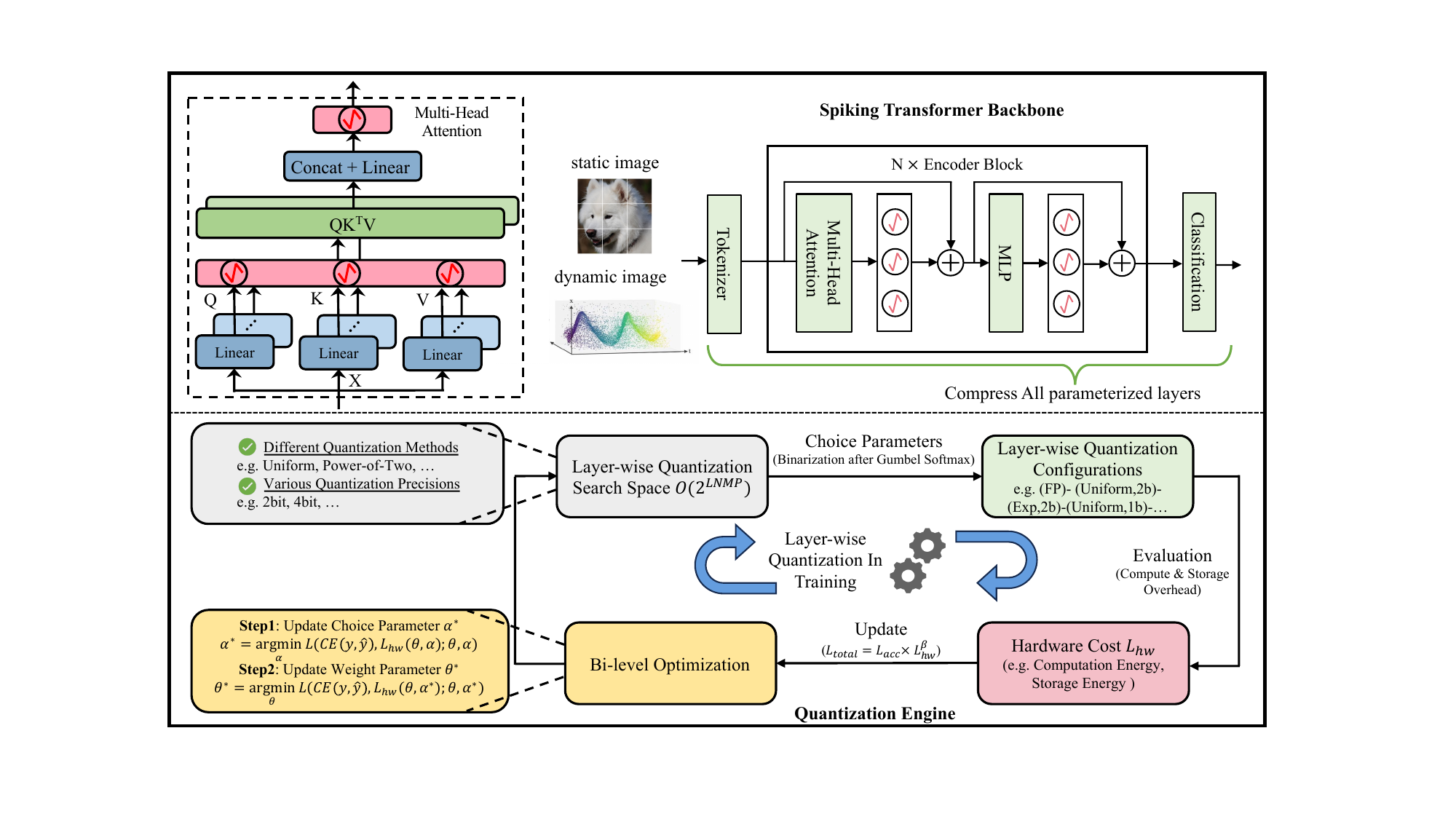}
    \caption{{\method}: proposed heterogeneous quantization by bi-level neural architecture search.}
    \label{fig:spikehq}
\end{figure*}

Spiking transformers are advantageous in their improved biological plausibility. Equally importantly, they can lead to energy efficient processing due to the binary nature of spiking activities. However, the last key benefit has not been fully exploited in the existing spiking transformer architectures. In \citep{spikformer}, even though the spiking neurons produce one-bit binary outputs, due to the residual connections across different blocks, the output of a block are multi-bit integers, which are inputs to the subsequent block. Furthermore, the parameters of  each spiking transformer layer are floating point numbers to maintain accuracy, rendering expensive multi-bit additions and multiplications and high storage overheads. 

We argue that the characteristics of spiking neural networks, when explored properly, would offer ample opportunities for weight quantization. Spiking activations are binary, signifying the robustness of spike-based computation with respect to parameter imprecision, and provides room for  weight quantization either uniformly or in other means. 
Multipliers are much more costly in terms of area, power, and latency compared with shifters and adders. 
Multiplications can be efficiently performed with addition and shift operations\citep{xue1986adaptive} with simple  hardware implementation at higher speeds \citep{marchesi1993fast, sanchez-romero2013approach}. On present-day processors, bit-shift instructions are faster than multiply/divide instructions and can be leveraged to perform multiplications and divisions by powers of two. Multiplication (or division) by a constant can be implemented using a sequence of shifts and adds (or subtracts). This would particularly favor the power-of-two quantization, which we explore along with uniform quantization in  {\method}.

%
\subsubsection{Uniform Quantizer}
Uniform quantization is defined by three parameters: the scale factor $s$, the zero-point $z$, and the bit-width $b$. $s$ and $z$ are both a floating-point number, and are used to map a floating-point value $\theta$ to an integer grid, whose size depends on $b$:
    \begin{equation}\label{eq:uniquant}
    \theta_{int}=\begin{cases}
       0, & \theta<0\\
       \lfloor\frac{\theta}{s}\rceil+z, & 0\leq \theta \leq 2^b-1 \\
       2^b-1, & \theta > 2^b-1
       \end{cases}
   \end{equation}
   \begin{equation}
       \theta\approx \hat{\theta}=s(\theta_{int}-z).
   \end{equation}

\subsubsection{Power-of-Two Quantizer}

Power-of-two quantization is symmetric with a power-of-two scale factor $s$. Scaling with $s$ can be efficiently realized by performing bit-shifting $k$ times. It is a suitable quantization choice for spiking transformers as the weight distributions are approximately symmetric. In spiking transformers such as \citep{spikformer}, leveraging the residual connection results in a progressive enlargement of spike activation range accumulation as the network depth increases. Consequently, the adoption of power-of-two quantization enables efficient shift operations over low-precision multipliers, also enhancing the compression rate.
The b-bit quantizer quantizes a floating point number $\theta$ into a power-of-two number as: 

\begin{equation}
s = 2^{\lfloor(log_2^{\text{max}(\theta)})\rfloor}
\end{equation}
\begin{equation}
\theta_{scaled} = \theta / s 
\end{equation}
\begin{equation}
\begin{aligned}
    \hat{\theta} = 
    \begin{cases}
        0 &\text{ if } -2^{-2^{b - 1} + 1 } < \theta_{scaled} < 2^{-2^{b - 1} + 1 } \\
        1 &\text{ if } \theta_{scaled} > 1\\
        -1 &\text{ if } \theta_{scaled} < -1\\
        -2^{\lfloor(log_2^{-\theta_{scaled}})\rfloor} &\text{ if } -1 < \theta_{scaled} < -2^{-2^{b - 1} + 1 }\\
        2^{\lfloor(log_2^{\theta_{scaled}})\rfloor} &\text{ if } 2^{-2^{b - 1} + 1 } < \theta_{scaled} < 1\\
    \end{cases}
    \end{aligned}
\end{equation}
\begin{equation}
    \theta_{quant} = \hat{\theta} \cdot s \label{eq:p2quant}
\end{equation}

The selection of the quantization scheme for each spiking transformer layer is determined by solving a neural network architecture search problem, as detailed in Section~\ref{sec:nas}.

\section{Heterogeneous Quantization by Neural Architecture Search}\label{sec:nas}

\subsection{Modeling of Hardware Overhead}
The proposed neural architecture search (NAS) finds the optimal heterogeneous quantizations for a given spiking transformer while jointly optimizing model accuracy and hardware overhead. We adopt the standard cross-entropy loss to evaluate model accuracy while  proposing an analytic model to evaluate hardware cost.

While an unquantized spiking transformer may involve float-point multiplications \citep{spikformer}, integer multipliers and shifters are used for realizing multiplications in layers with uniform and power-of-two quantization, respectively. We consider computation and memory access overhead, and model the hardware overhead for each layer under two different quantizers as follows targeting the 45nm CMOS technology: 
\begin{equation}\label{eqn_layerenergy}
    C_{HW}=\begin{cases}
        \#ops \times (C_{mult}+C_{add}) + \#bits\times C_{dram} & \text{if uniform}\\
        \#ops \times (C_{shift}+C_{add}) + \#bits\times C_{dram}& \text{if power of two}
    \end{cases}
\end{equation}
where $\#ops$ is the number of multiplications performed, $\#bits$ the storage in terms of number of bits, and $C_{shift}$, $C_{add}$, $C_{mult}$, $C_{dram}$ the energy cost per shift/add/mult/DRAM operation based on  Table~\ref{tab:energycost}. Specifically, $\#ops$ and $\#bits$ are given by:
\begin{equation}
    \#ops=\begin{cases}
    f_i\times f_o\times d & \text{for linear layer}\\
    k^2\times c_i\times c_o\times s  & \text{for conv layer}
    \end{cases}
\end{equation}
\begin{equation}
    \#bits=\begin{cases}
    f_i\times f_o\times b & \text{for linear layer}\\
    k^2\times c_i\times c_o\times b  & \text{for conv layer}
    \end{cases}
\end{equation}
where $f_i$ and $f_o$ are the input feature and output feature dimensions of the linear layer,  $d$ the input tensor dimension, $k$ the kernel size,  $c_i$ and $c_o$  the input and output channel size of the convolution layer respectively, $s$ the input image size, and $b$ the bit width of the quantization scheme. 

\begin{table}[h]
\centering 
\caption{Energy Cost (pJ) of Operations \citep{ShiftAddNet} }
\begin{tabular}{c|c|c|c|c}
\hline
\hline
Format\textbackslash{}OPs & Mult & Add  & Shift & DRAM \\ \hline
FP32 & 3.7  & 1.1  & 0.13  & 650  \\ \hline
INT8 & 0.2  & 0.03 & 0.024 & 163 \\
\hline
\end{tabular}
\label{tab:energycost}
\end{table}


\subsection{Problem Formulation}
For a spiking transformer with $N$ encoder blocks, $L$ layers in each transformer block, and $M$ candidate quantizers each with $P$ different quantization precision, we aim to find the optimal quantization scheme for each layer while jointly optimizing model accuracy and hardware overhead. Formally, we formulate this as a bi-level neural architecture search (NAS) problem: 
\begin{eqnarray}\label{eq:bi_opt}
&& \quad\min_{\bm q} \mathcal{L} \left( \bm q, \bm \theta^*\left(\bm q\right)\right) \label{top_prob} \\
&&  s.t. \quad \bm \theta^*\left(\bm q \right) = \argmin_{\bm \theta} \mathcal{L}\ \left(\bm q, \bm \theta\right) \label{bottom_prob},  
\end{eqnarray}
where $\bm \theta$ represents the weight parameter of the model, $\bm q$ is the architectural parameter and encodes the optimal quantization schemes including quantizer type and bit width for all layers, and $\mathcal{L}$ captures both model accuracy and hardware overhead. {\method} is illustrated in Fig.~\ref{fig:spikehq}.

\noindent \textbf{Quantization-aware training.} It is worth noting that the optimal model weight parameter $\bm \theta^*$ is obtained by solving the bottom-level optimization problem 
of (\ref{bottom_prob}), and is dependent on $\bm q$. This dependency gives rise to the bi-level nature of the NAS problem, and is due to the consideration of impact of quantization while optimizing weight parameter $\bm \theta$ through quantization-aware training. While a simpler strategy is to decouple quantization and weight parameter training by adopting post-training quantization, which however can render significant performance drop particularly with low-bit quantization. In contrast, quantization-aware training accounts for and adapts to the error caused by quantization and achieves better performance \citep{nagel2021whitequant}. Incorporating quantization-aware training in the solution to (\ref{bottom_prob}) requires back-propagating through the quantization effects reflected in (\ref{eq:uniquant})-(\ref{eq:p2quant}). We approximate the gradient using the straight-through estimator\citep{STE}, which approximates the gradient of the rounding operator as 1. During backward propagation, we use full-precision floating-point gradients for all quantities. We employ the straight-through estimator to approximate the derivative for the non-differentiable rounding functions employed in the uniform and power-of-two quantizations.
%
%
\subsection{Differential Architecture Search}
The architectural (quantization) parameter $\bm q$ in our NAS problem is discrete, giving rise to a large discrete solution space of $O({(MP)}^{LN})$ complexity. We adapt an efficient   differentiable neural architecture search approach \citep{liu2018darts} to solve the problem. 



\begin{figure*}[ht] 
    \centering
    \includegraphics[width=\textwidth, clip, trim={0.4cm, 0.3cm, 0.8cm, 0.3cm}]{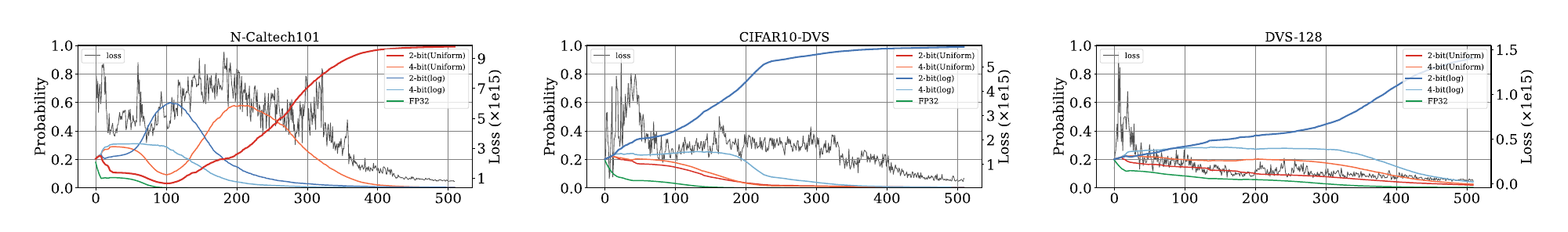}
    \caption{Evolution of quantization scheme selection probabilities for the last layer of spiking transformers across neuromorphic datasets and the total loss during architecture search.}
    \label{fig:blockchoice}
\end{figure*}

\noindent \textbf{Continuous-valued relaxation.} For each layer $l$ in block $j$, we introduce a continuous-valued selection probability parameter $\alpha_{li}^{<j>}$ for each $i$-th quantization scheme candidate in the set of all $MP$ options. 
We construct a composite layer presentation by summing up all possible quantized realizations of the layer weighted by the corresponding selection probability parameter $\alpha_{li}^{<j>}$:
\begin{equation} \label{eq:forward}
        y_l = \sum_{i=1}^{MP} g(\alpha_{li}^{<j>}) \cdot f_{\theta_{li}^{<j>}} (y_{l-1}),
\end{equation}
where $g(\cdot)$ represents a suitable form of Gumbel softmax \citep{gumbelsoftmax}, $\theta_{li}^{<j>}$ is the weight parameter under quantization scheme $i$, $f_{\theta_{li}^{<j>}}(\cdot)$ maps the output of layer $l-1$, i.e. the input to layer $l$, denoted by $y_{l-1}$, to the output of layer $l$'s output $y_l$.  
We then construct a ``supernet'' for the entire transformer by cascading all composite layers. 

We create a continuous-valued relaxation to the NAS problem of (\ref{top_prob}, \ref{bottom_prob}), optimizing the supernet over the continuous-valued weight parameter $\bm \theta$ and all $\alpha_{li}^{<j>}$. Once this relaxed bi-level problem is solved, the optimal quantization scheme for each layer $i$ is chosen by the option that has the largest selection probability $\alpha_{li}^{<j>}$.

The total hardware loss of the supernet is obtained by summing up all layer-wise hardware costs based on (\ref{eqn_layerenergy}) weighted by the corresponding quantization selection probabilities:  
\begin{equation}\label{softcost}
    L_{hw} = \sum_{j=1}^{N}\sum_{l=1}^L \sum_{i=1}^{MP} g(\alpha_{li}^{<j>})  C_{HW,li}^{<j>}. 
\end{equation}

We define the total loss to be minimized based on  a combination of cross-entropy model accuracy loss $L_{acc}$ and hardware loss $L_{hw}$ with a user-specified $\beta$ trading off between the two:
\begin{equation}\label{loss}
    \mathcal{L}_{total}  =  L_{acc}\times L_{hw}^{\beta}.
\end{equation}


\noindent \textbf{Solving the relaxed bi-level optimization problem.} 
To solve the relaxed bi-level problem efficiently, we adopt an iterative two-step approach. In Step 1, we update  all $\alpha_{li}^{<j>}$,  and then in Step 2, we update $\bm \theta$, both using backpropagation.

The standard Gumbel softmax is given by: 
\begin{equation}
\label{eq:gsoftmax}
     GS(\alpha_i) =\frac{\text{exp}((\text{log}\alpha_i + G_i)/\lambda)}{\sum_{i=1}^{MP}\text{exp}((\text{log}\alpha_i + G_i)/\lambda)} 
\end{equation}
where the temperature hyperparameter $\lambda \in (0, \infty)$ , $G_i$ is an i.i.d random variable sampled from the Gumbel(0,1) distribution. $G_i$ adds random noise into the standard softmax.
The smaller the $\lambda$, the closer the Gumbel softmax $GS(\cdot)$ is to the typical softmax.
For both updating steps, we initially set  $\lambda$ to be high to encourage exploration of diverse quantization choices at the beginning, and then gradually reduce $\lambda$ to stabilize the optimization towards  one specific choice per layer. While the standard Gumbel softmax $GS(\cdot)$ is used in Step 2 for the weight parameter, the one-hot version of it is adopted in Step 1 to update the quantization selection probabilities: 
\begin{equation} \label{gfunc}
    g(\alpha_k)= \begin{cases}
        GS(\alpha_k)& \text{when updating $\bm \theta$} \\
        \text{One-hot}(GS(\alpha_k))& \text{when updating $\alpha_{li}^{<j>}$} 
    \end{cases}
\end{equation}
Essentially, the use of $GS(\cdot)$ in Step 2 allows the weight parameters under all quantization choices to be updated in one step. In Step 1, we turn the selection probabilities into a single one-hot quantization scheme and only update the selection probability weights under this quantization choice.  This strategy helps speed up the convergence\citep{autonba}.

\section{Experiments}
\subsection{Experimental Setup}

\noindent \textbf{Datasets/models.} We evaluated the proposed heterogeneous quantization on spiking transformer models trained on three widely adopted neuromorphic datasets: DVS128, N-Caltech101\citep{ncaltech101}, and CIFAR10DVS \citep{cifar10dvs}. We followed the same model architectures and used the full precision models with 32-bit floating point weights from \citep{spikformer} as reference. DVS128 contains 11 hand gesture categories from 29 individuals under 3 illumination conditions. The Neuromorphic Caltech101(N-Caltech101) dataset is a spiking version of the frame-based Caltech101 dataset \citep{ncaltech101}. It comprises a diverse collection of images categorized into 101 distinct object classes, ranging from animals and vehicles to household items and everyday objects. CIFAR10-DVS is an event-stream dataset for object classification \citep{cifar10dvs}. It was created by converting 10,000 frame-based images from the CIFAR-10 dataset into 10,000 event streams with sophisticated spatio-temporal structures. We executed architecture search by training a supernet packing different quantized spiking transformers with a 4 CONV-layer spiking tokenizer, two transformer blocks with 256-embedding features and 16-head spiking self-attentions (SSA) over 16 time steps.  We used the AdamW optimizer on 510 training epochs with the learning rate warmed up from 1e-4 initially to the maximum learning rate of 1e-3, then reduced with cosine decay down to the minimum learning rate of 1e-5. 

\noindent \textbf{Layer-wise quantization.}
Our quantization neural architecture search space encompassed 5 configurations: 32-bit floating point representation, 2-bit power-of-two quantization, 4-bit power-of-two quantization, 2-bit uniform quantization, and 4-bit uniform quantization, applied to every convolution or linear layer.




\begin{figure*}[ht]
    \centering
    \includegraphics[width=\textwidth,clip, trim={5cm 2cm 5cm 1cm}]{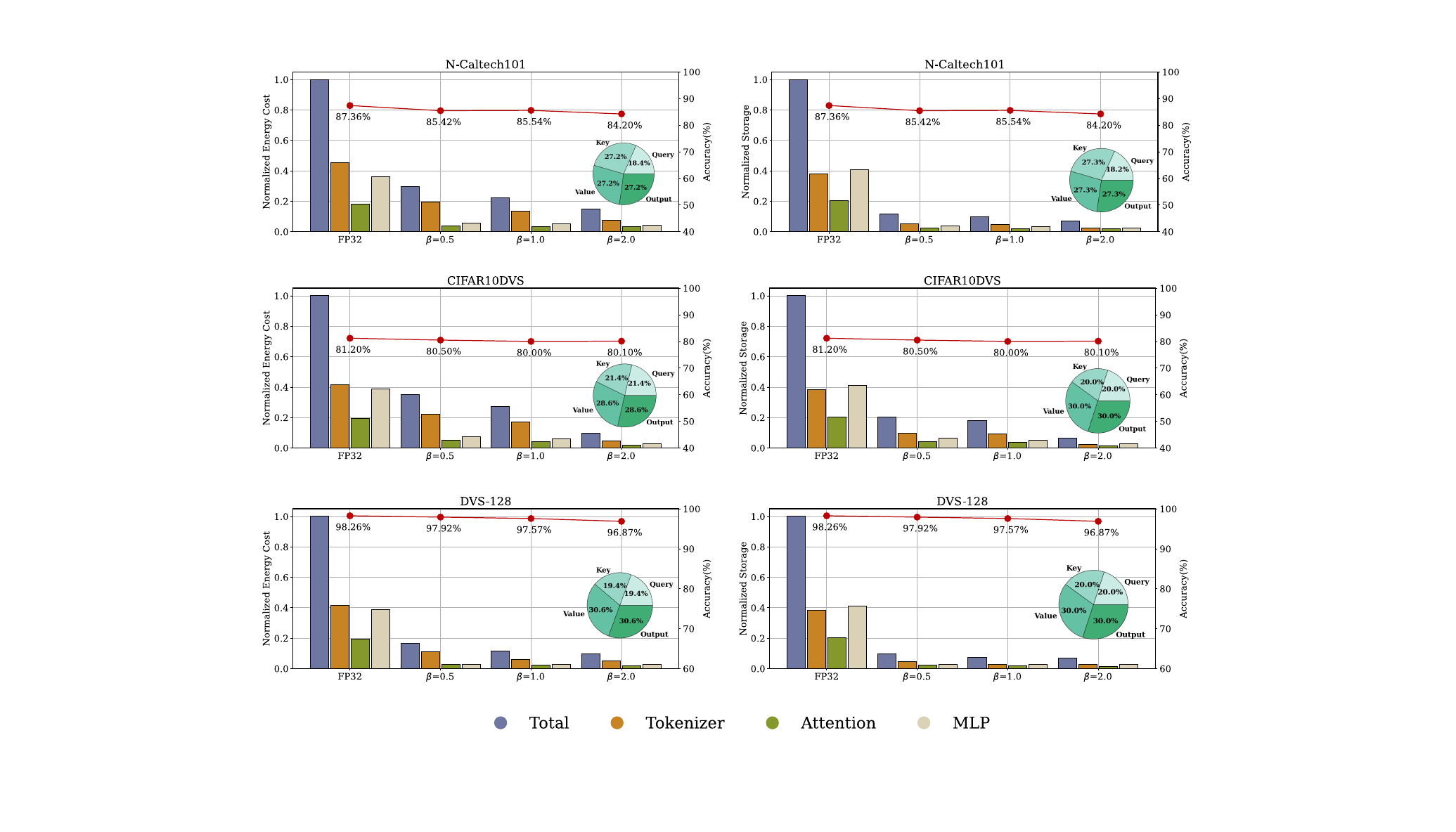}
    \caption{Breakdown of normalized energy consumption and storage overhead of spiking transformers across different neuromorphic vision tasks. The energy consumption and storage overhead are quantized across tokenizers, self-attention layers, and MLP layers before and after applying the proposed quantization method {\method}, under different hyperparameters. Red colored lines represent model accuracies; Pie charts break the energy consumption and storage of self-attention layers into the query, key, value, and output layers when $\beta=2.0$.}
    \label{fig:energystorage}
\end{figure*}
\begin{figure*}[ht]
    \centering
    \includegraphics[width=\textwidth, clip, trim={5cm 2cm 5cm 1cm}]{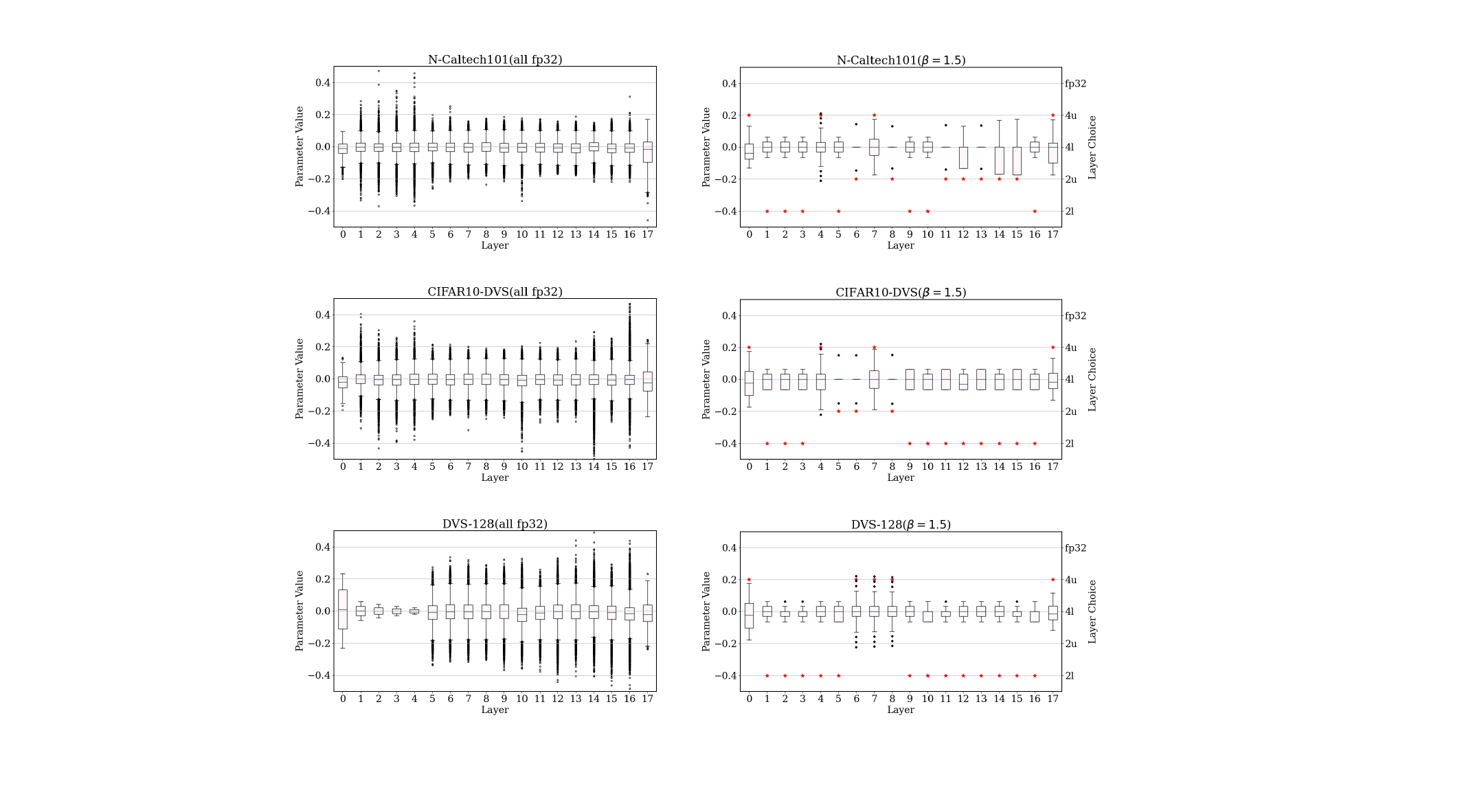}
    \caption{Distribution of Weight parameter value before(top figures) and after(bottom figures) applying {\method}. Median values are highlighted by blue lines, with boxes delineating the 25th and 75th percentile ranges. The external black lines represent the 5th and 95th percentiles, while outliers beyond these markers are denoted with black dots. After applying {\method}, the value range becomes more compact across different datasets; Red dots represent the layer-wise quantization choices: `2l' and `4l' for 2-bit and 4-bit power-of-two; `2u' and `4u' for 2-bit and 4-bit uniform; `fp32' for 32-bit floating-point.}
    \label{fig:param}
\end{figure*}

\subsection{Results on neuromorphic datasets}

\noindent \textbf{Accuracies, storage and energy costs.}
The effectiveness of our approach is clearly illustrated in Table~\ref{tab:results}, where $\beta$ represents the weighting hyparameter for hardware cost in the overall optimization objective. A zero-valued $\beta$ signifies the full-precision model. Across the range of $\beta$, {\method} results in only a marginal drop in accuracy, typically up to ~1\% on the DVS 128 and CIFAR10DVS datasets. {\method} significantly reduces the energy and storage requirements of the full precision models to as low as 9.73\% (10.42$\times$) and 6.58\% (15.20$\times$), respectively. The average number of bits used for weight parameters is reduced from 32 to 2.16bits.


\begin{table}[h]
\centering
\caption{Results of the heterogeneous quantization {\method}.}
\label{tab:results}
\begin{tabular}{c|c|c|c|c}

\hline
\multirow{2}{*}{Dataset} & \multirow{2}{*}{$\beta$} & \multirow{2}{*}{\shortstack{Accuracy \\ (\%/diff)}} & \multirow{2}{*}{\shortstack{Storage \\ (MB/\%/avg bits)}} & \multirow{2}{*}{\shortstack{Energy \\ (mJ/\%)}} \\
& & & & \\
\hline \hline
\multirow{5}{*}{DVS128}            & 0 & 98.26 / 0.00      & 10.21 / 100 / 32     & 14.99 / 100    \\ \cline{2-5} 
             & 0.5  & 97.90 / -0.34     & 0.982 / 9.61 / 3.07  & 2.508 / 16.73  \\ \cline{2-5} 
      & 1.0  & 97.57 / -0.69     & 0.744 / 7.28 / 2.33  & 1.741 / 11.61  \\ \cline{2-5} 
             & 1.5  & 97.22 / -1.04     & 0.688 / 6.74 / 2.16  & 1.471 / 9.81  \\ \cline{2-5} 
             & 2.0  & 96.87 / -1.39     & \textbf{0.695} / \textbf{6.80} / \textbf{2.18}  & 1.470 / 9.80   \\ \hline \hline
\multirow{5}{*}{\shortstack{CIFAR10 \\ DVS}}             & 0 & 81.20 / 0.00      & 10.21 / 100 / 32     & 14.99 / 100   \\ \cline{2-5} 
             & 0.5  & 80.52 / -0.70     & 1.033 / 10.11 / 3.24 & 2.615 / 17.44  \\ \cline{2-5} 
  & 1.0  & 80.00 / -1.20     & 0.909 / 8.90 / 2.85  & 2.583 / 13.60  \\ \cline{2-5} 
      & 1.5  & 80.40 / -0.80     & 0.802 / 7.86 / 2.52  & 2.039 / 11.48  \\ \cline{2-5} 
             & 2.0  & 80.10 / -1.10     & \textbf{0.672} / \textbf{6.58} / \textbf{2.11}  & 1.765 / 9.73   \\ \hline \hline
\multirow{5}{*}{\shortstack{Neuro- \\ Caltech \\ 101}}    & 0 & 87.36 / 0.00      & 10.30 / 100 / 32     & 17.49 / 100    \\ \cline{2-5} 
             & 0.5  & 85.42 / -1.94     & 1.222 / 11.85 / 3.79 & 5.157 /  29.49 \\ \cline{2-5} 
 & 1.0  & 85.54 / -1.82     & 1.010 / 9.8 / 3.13   & 3.918 / 22.40  \\ \cline{2-5} 
        & 1.5  & 85.30 / -2.06     & 0.814 / 7.9 / 2.53   & 3.073 / 17.57  \\ \cline{2-5} 
             & 2.0  & 84.20 / -3.15     & \textbf{0.704} / \textbf{6.84} / \textbf{2.19 } & 2.621 / 14.98  \\ \hline
\end{tabular}
\end{table}

\noindent \textbf{Evolution of quantization architecture search.} Figure \ref{fig:blockchoice} illustrates the architectural evolution in the quantization search for a query linear layer, designed for spiking self-attention, in the second block of each of the three transformers. At the initiation of the search, all five quantization configurations begin with an equal section probability of $20\%$. As the search progresses, one particular configuration is ultimately chosen. Notably, with an increasing dataset complexity from left to right (DVS 128, CIFAR10DVS, and N-Caltech101), the curves representing the selection probabilities become more intertwined, and the overall loss exhibits greater fluctuations. Each significant shift in the selection probabilities results in substantial changes in the loss—either a distinct increase or decrease—emphasizing the significant impact of the chosen quantization scheme on the overall loss. As the oscillations in the search process diminish, an optimized quantization choice emerges, leading to the attainment of the lowest achievable loss.


\noindent \textbf{Breakdown of energy and storage overheads.} In Figure \ref{fig:energystorage}, we break the total hardware overhead into three distinct components of the network architecture: tokenizer layers, spiking attention layers, and MLP layers. In the absence of quantization, full-precision models incur significant energy consumption and storage costs, particularly in the tokenizer and MLP layers. Taking the CIFAR10DVS dataset as an example, 41.4\% of the total energy is expended in the tokenizer, 19.5\% in self-attention, and 38.9\% in the MLP. Similarly, 38\% of the total storage is attributed to the tokenizer, 20\% to self-attention, and 41\% to the MLP. 
Following quantization, all layers experience substantial compression, leading to reduced storage costs. The tokenizer layer remains the predominant contributor to energy consumption. With a quantization parameter of $\beta=2.0$, the total energy is only 9.7\% of that of the full-precision model, distributed as 48\%, 20\%, and 30\% across the tokenizer, spiking attention, and the MLP, respectively. The total storage reduces to only 6.6\% of that of the full-precision model, with storage distribution among the tokenizer, spiking attention, and MLP at 36\%, 24\%, and 39\%, respectively. This analysis reveals that the gradual conversion of images into embedding patches requires substantial resources. Additionally, the energy of the spiking attention and MLP layers can be significantly reduced with a small $\beta$ value, such as 0.5, without compromising task performance. However, achieving a substantial energy reduction in the tokenizer necessitates a higher $\beta$ value, potentially leading to some performance loss. This underscores the tokenizer layer as the key bottleneck in the energy/performance tradeoff, suggesting that it should be the focus of architectural and design innovations.

\noindent \textbf{Disparity in layer-wise quantization.}
Figure \ref{fig:param} shows the distributions of weight parameters within spiking transformers before and after applying the proposed quantization method, and the optimized layer-wise quantization choices. The application of quantization enables a discrete and compact parameter value range, and clearly improves the sparsity by creating more centralized distributions. But, do different layers favor a similar quantization choice? 
Within the spiking self-attention layers, computing the query, key, value, and output is based on the same operations with the equal amount of parameters. This symmetry seems to suggest an equal importance of parameter precision for these four types of computations. Intriguingly, we quantize the post-compression energy and storage share for the four parts in the pie charts of Figure \ref{fig:energystorage}, which consistently shows that computing the query and key requires a lower parameter precision, which is the case even though the neural activities associated with the query and key tend to fire at lower rates than those with value and outputs\citep{spikformer}. Additionally, as shown in Figure \ref{fig:param}, the first and the last layer, which are the first convolution layer and the classification head, respectively, consistently necessitate 4-bit uniform quantization, which is the most high-precision and costly choice, signifying the criticality of these layers in the overall task performance.

\section{Conclusion}
This paper introduces a heterogeneous quantization approach tailored for emerging spiking transformers. The proposed {\method} strategically eliminates the heavy parameter challenge within spiking transformers and explores the optimal quantization scheme on a layer-by-layer basis,  optimizing overall energy and storage overhead, and model accuracy through neural architectural search. Our experimental studies underscore the potential of {\method}, demonstrating its capability to achieve up to an order of magnitude reduction in energy and storage requirements without significantly compromising model accuracy. Furthermore, our empirical findings highlight specific layers that require high-precision weights, signifying them as the focal points for future architectural and design innovations.

\nocite{langley00}

\bibliography{main}
\bibliographystyle{tmlr}

\end{document}